%% file: anonymous-submission-latex-2026.tex
\documentclass[letterpaper]{article} 
\usepackage{aaai2026}
\usepackage{times}  
\usepackage{helvet}  
\usepackage{courier}  
\usepackage[hyphens]{url}  
\usepackage{graphicx} 
\usepackage{natbib}  
\usepackage{caption} 
\usepackage{algorithm}
\usepackage{algorithmic}

\usepackage{newfloat}
\usepackage{listings}
\DeclareCaptionStyle{ruled}{labelfont=normalfont,labelsep=colon,strut=off} 
\floatstyle{ruled}
\newfloat{listing}{tb}{lst}{}
\floatname{listing}{Listing}
\nocopyright 

\title{Point-to-Point: Sparse Motion Guidance for Controllable Video Editing}
\author{
    Yeji Song, Jaehyun Lee, Mijin Koo, JunHoo Lee, Nojun Kwak
}
\affiliations{

    Seoul National University\\
    {\rm \{ldynx,jhlee795,starmj09,mrjunoo,nojunk\}@snu.ac.kr}
}

\input{shortcuts}

\begin{document}

\maketitle

\begin{abstract}
Accurately preserving motion while editing a subject remains a core challenge in video editing tasks. Existing methods often face a trade-off between edit and motion fidelity, as they rely on motion representations that are either overfitted to the layout or only implicitly defined. To overcome this limitation, we revisit point-based motion representation. However, identifying meaningful points remains challenging without human input, especially across diverse video scenarios. To address this, we propose a novel motion representation, anchor tokens, that capture the most essential motion patterns by leveraging the rich prior of a video diffusion model. Anchor tokens encode video dynamics compactly through a small number of informative point trajectories and can be flexibly relocated to align with new subjects. This allows our method, \methodname{}, to generalize across diverse scenarios. Extensive experiments demonstrate that anchor tokens lead to more controllable and semantically aligned video edits, achieving superior performance in terms of edit and motion fidelity.
\end{abstract}


\input{sections/01_introduction}
\input{sections/02_related_work}
\input{sections/03_method}
\input{sections/04_experiments}
\input{sections/05_conclusion}

\pagebreak
\bibliography{aaai2026}



\clearpage
\input{sections/06_appendix}
\clearpage


\end{document}

%% file: shortcuts.tex
\usepackage{amsfonts}       
\usepackage{amsmath}
\usepackage{booktabs}       
\usepackage{enumitem}
\usepackage[T1]{fontenc}    
\usepackage{graphicx}
\usepackage[utf8]{inputenc} 
\usepackage{makecell}
\usepackage{microtype}      
\usepackage{multirow}
\usepackage{nicefrac}       
\usepackage{tabularx}
\usepackage{adjustbox}
\usepackage{url}            
\usepackage[dvipsnames]{xcolor}
\usepackage[table,xcdraw]{xcolor}

\usepackage{longtable}
\usepackage{filecontents,catchfile,pgffor}

\usepackage{environ}

\newcommand*{\methodname}[0]{\textsc{Point-to-Point}}

\newcolumntype{Y}{>{\centering\arraybackslash}X}

\newcommand{\ie}{\textit{i}.\textit{e}.}
\newcommand{\eg}{\textit{e}.\textit{g}.}

\DeclareMathOperator*{\argmax}{argmax}
\DeclareMathOperator*{\argmin}{argmin}



%% file: sections/01_introduction.tex
\section{Introduction}

\begin{figure*}[ht!]
    \centering
    \includegraphics[width=\textwidth]{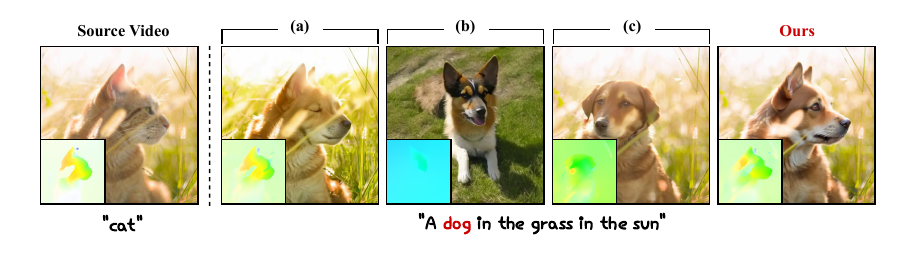}
    \vspace{-2\baselineskip}
    \caption{Comparison of editing results given a source video (\textbf{left}). The left-bottom boxes show the estimated optical flow~\cite{teed2020raft} from generated videos. \textbf{(a)} Signal-based method~\cite{cong2023flatten} produces overfitted layouts. \textbf{(b)} Adaptation-based method~\cite{zhao2024motiondirector} produces inaccurate motion. \textbf{(c)} Point-based method~\cite{gu2024videoswap} generates inaccurate motion when the points fail to capture meaningful motion. \textbf{Ours} successfully edits the subject while preserving the motion.} 
    \vspace{-1\baselineskip}
    \label{fig:problem}
\end{figure*}

\begin{figure*}[t!]
    \centering
    \includegraphics[width=\textwidth]{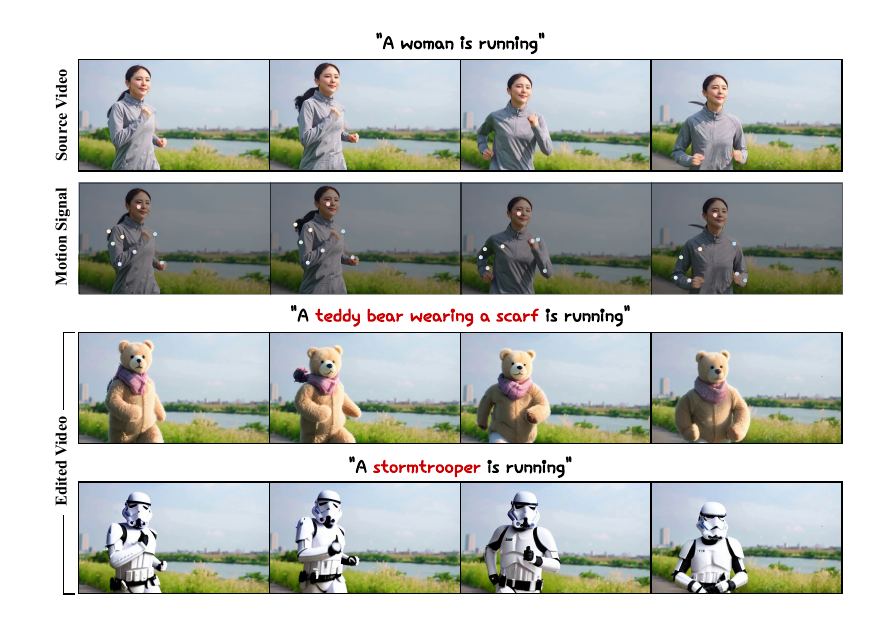}
    \vspace{-2\baselineskip}
    \caption{Given a source video, our \methodname{} extracts compact and essential motion information using point trajectories, and transfers it to guide edits across diverse target subjects, including humans, animals, objects, and even scene-level motion.} 
    \label{fig:teaser}
    \vspace{-1\baselineskip}    
\end{figure*}

Diffusion Models (DMs)~\cite{ho2020ddpm,song2020score,song2021ddim,rombach2022ldm} and Video Diffusion Models (VDMs)~\cite{ho2022video,singer2022make,ho2022imagenvideo,hong2022cogvideo,esser2023structure,blattmann2023align} have opened new possibilities to synthesize and edit videos with high visual fidelity. A key challenge in video editing is preserving the motion of a source video while modifying the moving subject. Video editing models first encode the motion in a source video through extracted signals or learned features, and then synthesize new videos based on those signals or features. Initial approaches have used explicit motion signals based on low-level structures such as optical flow, depth maps and feature correspondences~\cite{zhang2023controlvideo,cong2023flatten,geyer2023tokenflow,liang2024flowvid,wang2024cove,yang2025videograin}. Alternatively, other methods have modeled motion dynamics in the source video through adaptation, either by tuning network parameters~\cite{zhao2024motiondirector} or refining latent representations~\cite{yatim2024space,jeong2024dreammotion}. However, each of two main approaches comes with distinct limitations. As signal-based methods leverage dense signals (\ie, signals encoded at every pixel), they inherently conform to the source layout. When the new subject has a significantly different appearance, they often result in an awkward body structure or blurry videos, as shown in Fig.~\ref{fig:problem}(a). On the other hand, adaptation-based methods encode motion implicitly through optimization, therefore often struggle to accurately transfer the motion. As shown in Fig.~\ref{fig:problem}(b), the generated motion deviates noticeably from the source video.

Recently, several point‑based methods~\cite{gu2024videoswap,tu2025videoanydoor} have been proposed to leverage semantic points for video editing to remedy the shortcomings of signal-based approaches. Nevertheless, despite the clear benefits, this direction has been relatively underexplored due to difficulties in \textit{obtaining meaningful points}. A common workaround in existing methods has been to rely on manually annotated points or off-the-shelf pose estimation models. However, this approach does not fully solve the problem. Videos often include non-human subjects with diverse structures, and the subjects are frequently captured mid-action, appearing blurry and in uncommon poses. Therefore, requiring users to manually annotate semantic points or choose a suitable estimator for each video is impractical. For open-world estimators~\cite{shi2023matching,xu2022pose,zhang2023clamp,yang2024x}, users are still required to provide reference keypoints or carefully curated prompts. Also, they often provide points that are often not fully aligned with the motion, resulting in motion discrepancies and degraded quality. As shown in Fig.~\ref{fig:problem}(c), applying an open-world pose estimator\cite{yang2024x} to generate keypoints for the existing method~\cite{gu2024videoswap} results in misaligned motion with the source video.

To overcome these challenges, we propose novel, fully-automated and motion-aligned points called \textbf{anchor tokens}. Our intuition is that, by using pre-trained video diffusion models, which already capture rich semantic and temporal information, we can extract meaningful point representations.
First, we define motion tokens, which are a set of extracted features using a video diffusion model. In contrast to prior works~\cite{geyer2023tokenflow,wang2024cove}, where features (often referred to as ``tokens") are independently extracted from each latent pixel in every frame, we track those tokens across frames and obtain distinct motion trajectories to extract more representative features of motion dynamics.
From this set, we select a subset of anchor tokens that represent the most informative motion patterns in the video. Each anchor token captures a distinct local motion, and together they form a comprehensive summary of the overall motion dynamics.

Meanwhile, the new subject in the edited video may not align well with the anchor tokens extracted from the source video. To address this, we adjust the anchor tokens to better match the new layout by matching each anchor token to the motion tokens in the edited video whose feature is most similar. Then, we relocate each anchor token to its corresponding position. This naturally aligns anchor tokens with the new subject. Furthermore, since anchor tokens operate in the latent space, the relocated anchor tokens integrate seamlessly with the edited video. 

As shown in Fig.~\ref{fig:teaser}, our method successfully generates edited videos by using anchor tokens that are well aligned with the motion and readily adapt to the new subject. Extensive experiments show that our method using anchor tokens outperforms existing approaches in terms of editing flexibility, motion consistency, and overall edit quality. Qualitative and quantitative results demonstrate that anchor tokens enable more coherent and semantically consistent edits across a wide range of video editing scenarios.

In summary, we introduce anchor tokens as a new motion guidance. They require no user effort, provide accurate motion information, and are adaptable enough to be readily applied to various subjects. By using anchor tokens, our method \methodname{} fully leverages the true advantages that motion points can offer to the model.

%% file: sections/02_related_work.tex
\section{Related Work}

\textbf{Video Generation and Video Editing.}
With the remarkable progress of diffusion models~\cite{ho2020ddpm,song2020score,song2021ddim,rombach2022ldm}, various methods have been proposed for video generation~\cite{ho2022video,singer2022make,ho2022imagenvideo,hong2022cogvideo,esser2023structure,blattmann2023align}. Recently, substantial efforts have been made to enhance the quality of the generated videos by conditioning the models on sketches, depth maps, camera poses, and bounding boxes~\cite{guo2024sparsectrl,wang2024motionctrl,ma2024trailblazer,jain2024peekaboo}.
Beyond video generation, there have  been growing research interest in video editing~\cite{wu2023tune,liu2024video,zhang2023controlvideo, cong2023flatten,wang2024cove,geyer2023tokenflow,zhao2024motiondirector,yatim2024space, jeong2024dreammotion,song2024save}, which aims to modify subject or content conditioning on a given source video. Many existing video editing approaches either incorporate explicit motion signals such as depth maps~\cite{zhang2023controlvideo}, optical flow~\cite{cong2023flatten}, points~\cite{gu2024videoswap,tu2025videoanydoor} or feature correspondences~\cite{wang2024cove,geyer2023tokenflow}, or learn motion information implicitly, optimizing additional weights or latent representations~\cite{zhao2024motiondirector,yatim2024space,jeong2024dreammotion}. In this paper, we extend point-based approach by introducing novel anchor tokens that offer precise and consistent motion guidance across diverse subjects.

\medskip
\noindent \textbf{Point-based Generation.}
Recently, point-based video generation~\cite{yin2023dragnuwa,wu2024draganything} has emerged as a promising topic. In this setting, the user specifies keypoints on the first frame along with their trajectories, and the model generates the remaining frames such that the motion follows these trajectories.
Several video editing methods~\cite{gu2024videoswap,tu2025videoanydoor} use point trajectories extracted from the source video to guide motion. Meanwhile, these methods typically require the user to explicitly provide points, which places a heavy burden on users. They also exhibit degraded editing quality when the selected points are inaccurate or misaligned with the motion. In contrast, our anchor tokens eliminate the need for user annotation and leverage the model’s own features, providing motion guidance that improves both motion alignment and practicality.

%% file: sections/03_method.tex
\section{\methodname{}}

\begin{figure*}[t!]
    \centering
    \includegraphics[width=\textwidth]{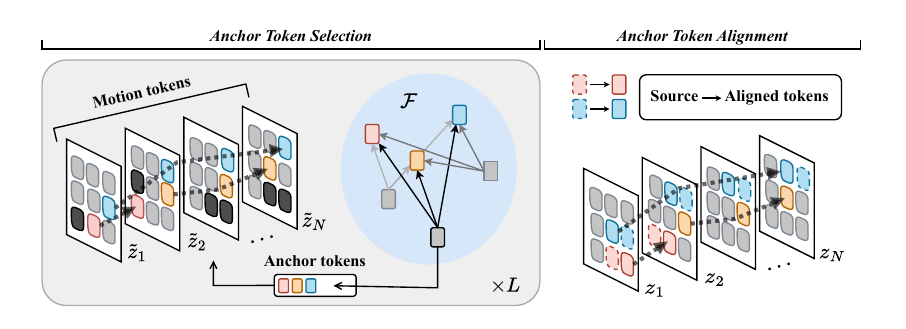}
    \vspace{-2em}
    \caption{\textbf{Left:} From motion tokens tracked across latent features $\tilde{z}_1, \dots, \tilde{z}_N$, we select a sparse set of anchor tokens (colored) that capture representative motion trajectories by computing similarity in feature space, denoted as $\mathcal{F}$, while filtering out redundant or out-of-subject tokens (black).
    \textbf{Right:} During editing, anchor tokens are aligned by identifying semantically corresponding locations in the edited video and injected at those positions, enabling motion transfer to the new subject.}
    \label{fig:method}
    \vspace{-1\baselineskip}
\end{figure*}

In this section, we propose \methodname{} that first uses fully-automated, model‑aligned \textit{anchor tokens} to capture representative motion trajectories. First, we define the problem we aim to solve. Then, we introduce our anchor token selection strategy, as well as explain how these tokens transfer motion. Finally, we describe how to align and adapt these points effectively to the new target for editing. The overview of our method is illustrated in Fig.~\ref{fig:method}.

\subsection{Problem Definition} \label{subsec:problem}

Given a source video \(\tilde{X}=\{\tilde{x}_1 , \tilde{x}_2, \dots \tilde{x}_N\}\) where \(\tilde{x}_i\) is the frame at time \(i\), and a conditioning prompt \(\mathcal{C}\) specifying the desired edit, the objective of video editing task is to generate an edited video \(X=\{x_1, x_2, \dots x_N\}\). For successful video editing, the following capabilities are required:

\begin{itemize}[leftmargin=*,noitemsep,topsep=0em]
    \item \textbf{Edit fidelity}: The edits specified by \(\mathcal{C}\) should be faithfully realized in each frame of \(X\).
    \item \textbf{Motion fidelity}: The original motion of \(\tilde{X}\) must be preserved in \(X\) wherever it does not conflict with \(\mathcal{C}\).
    \item \textbf{Practicality}: Motion transfer from \(\tilde{X}\) to \(X\) should be performed automatically, requiring minimal human effort.
\end{itemize}

To meet these requirements, the generative process often incorporates additional motion guidance from the source video \(\tilde{X}\). Signal-based methods~\cite{zhang2023controlvideo,geyer2023tokenflow,cong2023flatten,wang2024cove,liang2024flowvid,yang2025videograin} extract and leverage an explicit motion signal from the source video, whereas adaptation-based methods~\cite{zhao2024motiondirector,yatim2024space,jeong2024dreammotion} embed motion implicitly in the model or latent space, optimizing those representations on the source video. Point-based methods~\cite{gu2024videoswap,tu2025videoanydoor} guide the motion using semantic points and their trajectory. Despite their respective strengths, all three approaches struggle to satisfy the requirements above. Signal-based methods overfit to the spatial layout of the source video, often compromising edit fidelity. Adaptation-based methods often struggle to capture precise motion dynamics, leading to low motion fidelity. Point‑based methods require manual annotations, limiting their practicality. Furthermore, they can result in degraded edit and motion fidelity when points are inaccurate. Therefore, our goal is to leverage points that transfer precise, layout‑adaptive motion without manual annotations.

First, we adopt the pre-trained video diffusion model and extract the set of motion tokens \(\mathcal{M} = \{ m_1, m_2, \cdots, m_K \}\) from the source video where \(K\) is the total number of motion tokens. Each token \(m_k\) is associated with a spatiotemporal trajectory \(\phi_k \in \Phi\). Each trajectory is defined as a sequence of spatial coordinates over time:
\begin{align}
\phi_k = \left\{(u_1^k, v_1^k), \cdots, (u_N^k, v_N^k) \right\}
\label{eq:trajectory}
\end{align}
where $\left( u_i^k, v_i^k \right)$ indicates the location of token $m_k$ at frame $i$. 
We then define a subset of $L$ representative anchor tokens, denoted as $\mathcal{A} = \{ m_k \mid k \in \mathcal{I}_L \} \subset \mathcal{M}$, where $\mathcal{I}_L \subset \{1, \cdots, K\}$ is the index set and $|\mathcal{I}_L| = L < K$. Each anchor token $m_k \in \mathcal{A}$ is selected such that its trajectory $\phi_k$ captures a unique local motion pattern, serving as a sparse representation of the overall motion. 

\subsection{Anchor Token Selection} \label{subsec:selection} 

Selecting meaningful anchor tokens from a video has several fundamental challenges. 
First, it requires consistently identifying points on moving objects and reliably tracking their full trajectories, even through occlusions or when they move out-of-frame. Then, from the resulting trajectories, we have to select a salient subset that truly represents the video’s core dynamics.
This section details our approach to addressing these challenges. 

\medskip
\noindent \textbf{Trajectory Collection.} We begin by computing bidirectional optical flow~\cite{teed2020raft} between source frames $\{\tilde{x}_1 , \tilde{x}_2, \cdots, \tilde{x}_N\}$ and downsample the flow fields to match the resolution of the latent space.
In real-world videos, objects may become temporarily hidden (occluded) or move out of the camera frame, which often leads to incomplete trajectories that only appear in a part of the video. To better capture such motions, we sample three keyframes (\(\{1, \lfloor \frac{N}{2} \rfloor, N\}\)-th frames in our implementation) from the video. For each of these keyframes, we track how pixels change over time in the latent space and obtain their individual trajectories, collecting a richer set of trajectories that better cover the entire motion in the video. When the height and width of the latent feature map are denoted by $H$ and $W$, this yields a total of $3 \times H \times W$ initial trajectories. We then remove redundant trajectories and, when a subject mask is available, discard trajectories that fall outside the subject region. Finally, we obtain $K$ trajectories, denoted as $\Phi = \{ \phi_1, \phi_2, \cdots, \phi_K \}$.

\begin{figure*}[!htbp]
    \centering
    \includegraphics[width=\textwidth]{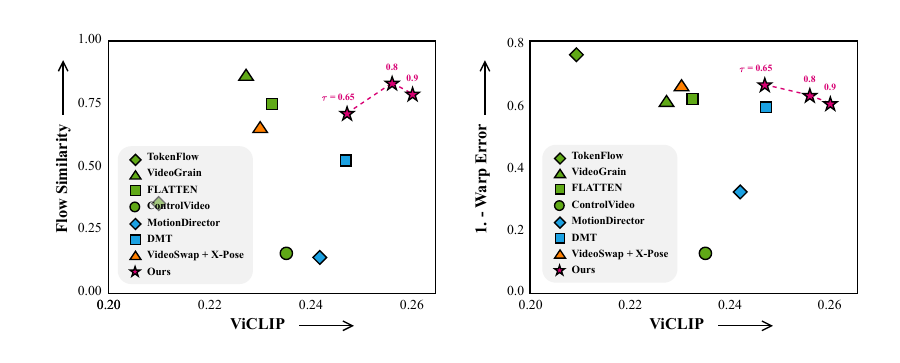}
    \vspace{-2em}
    \caption{Quantitative comparison on edit fidelity (x-axis) and motion fidelity (y-axis)}
    \label{fig:quantitative}
    \vspace{-1\baselineskip}
\end{figure*}

\medskip
\noindent \textbf{Anchor Token Selection.}
Along the trajectories, we extract the corresponding video diffusion features from intermediate blocks of the model. Following~\cite{xiao2024video}, we use normalized features, which have been shown to capture rich motion information. For each trajectory $\phi_k$, we obtain motion token \(m_k\) as:

\begin{align}
m_k = \big[\tilde{f}_1(u_1^k, v_1^k), \tilde{f}_2(u_2^k, v_2^k), \cdots \tilde{f}_N(u_N^k, v_N^k) \big]^{\top}
\label{eq:feature_trajectory}
\end{align}

where $\tilde{f}_i(u_i^k, v_i^k)$ denotes the normalized feature at location $(u_i^k, v_i^k)$ in frame $i$ of the source video. 
We define the distance between $m_k$ and $m_{k'}$ as the sum of their frame‑wise inner products, meaning that more similar tokens (with higher inner product values) have smaller distances.
\begin{align}
d(m_k, m_{k'}) = 1 - \sum^N_{i=1} \tilde{f}_i(u_i^k, v_i^k)^{\top} \tilde{f}_i(u_i^{k'}, v_i^{k'})
\label{eq:distance}
\end{align}
Having computed pairwise feature distances, we aim to select a small set of anchor tokens such that each token represents a salient local motion and is maximally distinct from the others. To this end, we adopt Farthest Point Sampling (FPS), a greedy selection algorithm originally popularized in 3D point cloud processing~\cite{qi2017pointnet++,shi2019pointrcnn,yang20203dssd,zhang2022not}. FPS starts from an arbitrary seed and, at each step, adds the point whose minimum distance to the already selected set is maximal. 
In our setting, let $\mathcal{I}_l \subset \{1, \cdots, K\}$ denote the index set of the $l$ previously selected anchor tokens. The corresponding anchor token set is given by $\mathcal{A}_l = \{ m_k \mid k \in \mathcal{I}_l \}$. We then sample the next anchor index $k^* \notin \mathcal{I}_l$ based on a diversity criterion: 
\begin{align}
k^* = \argmax_{k \notin \mathcal{I}_l} \left( \min_{\;\; k' \in \mathcal{I}_l} d(m_k, m_{k'}) \right), \mathcal{I}_{l+1} = \mathcal{I}_l \cup \{ k^* \}.
\end{align}
This results in a set of anchor tokens $\mathcal{A} = \left\{ m_k \mid k \in \mathcal{I}_L \right\}$ that are both mutually distinct and representative of unique local motion. We also set $L$ adaptively for each video by stopping the sampling process when the distance between the newly selected token $m_{k^*}$ and any previously selected token falls below a threshold $\tau$, \ie, they are considered similar. 

After the selection, we inject motion information of each anchor token into the latent space during inference. Inspired by \cite{gu2024videoswap}, each anchor token is projected by small, learnable MLPs and added element‑wise along its trajectory to the intermediate feature maps. Anchor tokens are injected only at their locations, leaving other regions unchanged, which allows motion transfer to occur precisely where needed and stay aligned with the new editing layout.

\subsection{Anchor Token Alignment} \label{subsec:alignment}


In video editing task, new subjects often have significantly different body structures. In that case, even if anchor tokens are successfully identified in the source video, applying them directly to the new subject without adaptation can lead to misalignment. Prior methods attempt to address this by either requiring users to manually specify points on the new subject based on their own judgment~\cite{gu2024videoswap}, or by introducing a separate appearance encoder and incorporating it through cross-attention~\cite{tu2025videoanydoor}. Meanwhile, our anchor tokens can more readily adapt to new subjects without requiring user input or an additional encoder. Specifically, we extract features in the latent space along the trajectories of the motion tokens in the middle of the denoising process during inference.
Then, we collect motion tokens $\{m_1^{\text{tgt}}, m_2^{\text{tgt}}, \cdots, m_K^{\text{tgt}}\}$ from $f$ where $f$ denotes the normalized features of the generated video (in contrast to $\tilde{f}$, which denotes features from the source video). For each anchor token $m_k \in \mathcal{A}$, we identify its most similar counterpart $m^{\text{tgt}}_{j_k}$ from the target motion tokens by minimizing distance defined in Eq.~\ref{eq:distance}:
\begin{align}
j_k = \argmin_{j \in \{1, \cdots, K\}} \, d(m_k, m_j^{\text{tgt}}), \quad \forall m_k \in \mathcal{A}
\label{eq:anchor_matching}
\end{align}

Then, we inject its motion information along the trajectory $\phi_{j_k}$, rather than using its original trajectory $\phi_k$ for the remaining timesteps of the denoising process. Relocating anchor tokens to their matched target trajectories ensures that motion information is injected in spatially correct regions, resulting in more coherent motion guidance.


%% file: sections/04_experiments.tex
\begin{figure*}[!t]
    \centering
    \includegraphics[width=\textwidth]{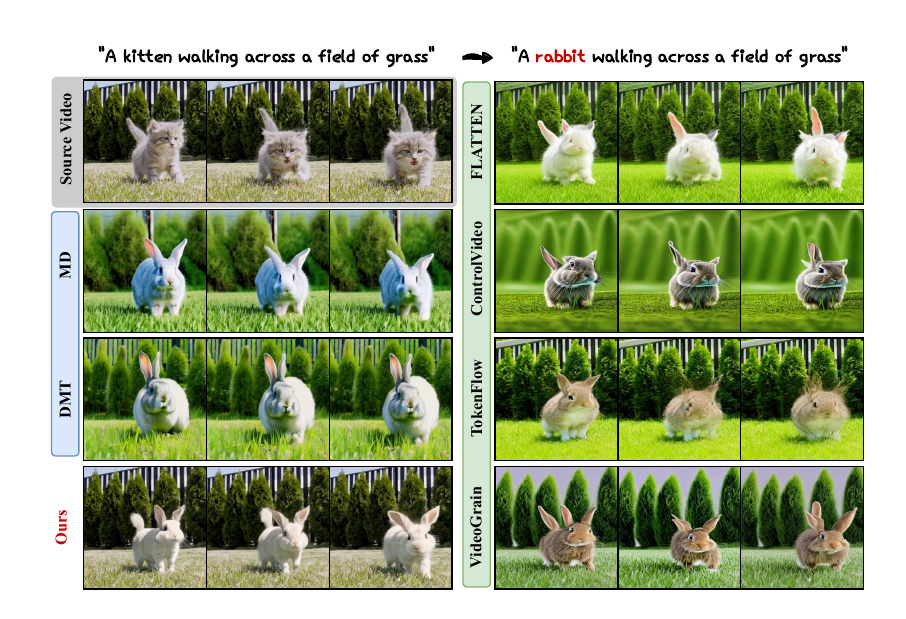}
    \vspace{-3em}
    \caption{Qualitative comparison of video editing results across various videos.
    }
    \label{fig:qualitative}
    \vspace{-1\baselineskip}
\end{figure*}

\section{Experiments}

We evaluated our method and baselines on general video editing tasks, revealing a clear trend across the methods. We further assess their performance in a more challenging scenario that demands both high edit and motion fidelity.

\subsection{Experimental Settings}

\textbf{Baselines.} We compare our method with state-of-the-art and concurrent works from three categories: signal-based, adaptation-based, and point-based methods. Signal-based methods, including ControlVideo~\cite{zhang2023controlvideo}, VideoGrain~\cite{yang2025videograin}, FLATTEN~\cite{cong2023flatten}, and TokenFlow~\cite{geyer2023tokenflow}, utilized depth maps, optical flow, and feature correspondences extracted from source video, respectively. These methods incorporate these signals either by adopting ControlNet~\cite{zhang2023controlnet} or by propagating them during inference. The adaptation-based baselines are MotionDirector (MD)~\cite{zhao2024motiondirector} and Diffusion Motion Transfer (DMT)~\cite{yatim2024space}, which optimized additional LoRA~\cite{hu2022lora} or latent representations on the source video to capture and transfer its motion. We also compare against VideoSwap~\cite{gu2024videoswap} where we use keypoints obtained from X-Pose, an open-world pose estimation model~\cite{yang2024x}. We provide the implementation details in the Appendix.

\medskip
\noindent \textbf{Evaluation.} We evaluate the quality of editing using three automatic metrics: ViCLIP Score, Flow Similarity, and Warp Error. For edit fidelity, we use ViCLIP Score~\cite{wang2023internvid,huang2024vbench}, which measures ViCLIP-based similarity with the input prompt. For motion fidelity, we use two metrics: Flow Similarity, which measures the cosine similarity between the optical flow fields of the source and edited videos, and Warp Error, which evaluates spatial consistency by computing pixel-wise differences after warping the edited frame using the source flow (scaled by $\times 10^{-4}$ for readability). For human evaluation, we conducted a user study using Amazon Mechanical Turk (AMT). A total of 51 participants evaluated videos, selecting the better results (ours vs.\ baseline) based on four criteria: (1) Edit Fidelity: how well the subject matches the reference or text prompt, (2) Motion Fidelity: how faithfully the motion reflects the source video, (3) Temporal Consistency: how smooth and stable the video appears, and (4) Overall Preference. 
Responses are normalized such that our method scores 1.0, with other methods reported as ratios, indicating relative preference compared to ours.
While Flow Similarity and Warp Error are sensitive to the source video’s layout since they rely on optical flow, Temporal Consistency offers a more perceptual measure that better aligns with human judgment.

\subsection{General Video Editing Evaluation}

In this experiment, we evaluate the general editing ability of our method and the baselines using 32 videos collected from LOVEU-TGVE dataset~\cite{wu2023cvpr}. Results are shown in Fig.~\ref{fig:qualitative} (qualitative) and Fig.~\ref{fig:quantitative} (quantitative). As shown in Fig.~\ref{fig:qualitative},
signal-based methods rely on dense motion signals and often generate the videos whose layout is constrained by the source video. Adaptation-based methods frequently suffer from motion inconsistency. Our method, based on sparse anchor tokens, provides layout-agnostic yet explicit motion guidance, mitigating both types of failure. This trend is also reflected quantitatively in Fig.~\ref{fig:quantitative}, where signal-based methods (colored green) better preserve motion but struggle with edit fidelity, whereas adaptation-based methods (colored blue) achieve higher edit fidelity at the cost of motion preservation.
ControlVideo, despite being signal-based, show low motion fidelity due to the unstable depth estimation in complex scenes. For VideoSwap, which uses X-Pose keypoints, the wide variety of video scenes in the dataset often results in inaccurate points, leading to degraded motion and edit fidelity. Our method advances the Pareto frontier by achieving strong results across both dimensions, edit and motion fidelity. More various results are also provided in the Appendix.

\subsection{Customized Subject Swapping} \label{subsec:customized}

In general video editing scenario, evaluation of edit fidelity can be hindered by the ambiguity of the text prompt. For instance, a prompt like "replace the dog with a cat" lacks detailed guidance, allowing the model to generate a cat that still conforms to the original dog’s layout. This makes it difficult to evaluate whether the model has truly captured the target identity or simply adapted it to the existing layout. In this regard, customized subject swapping~\cite{wei2024dreamvideo,gu2024videoswap,huang2025videomage} provides a more reliable setting for evaluation. Customized subject swapping aims to replace the subject in a source video with a user-specified identity while preserving the original motion. To evaluate this setting, we compare our method with FLATTEN~\cite{cong2023flatten}, TokenFlow~\cite{geyer2023tokenflow}, and MotionDirector~\cite{zhao2024motiondirector}, all of which support end-to-end subject customization via LoRA~\cite{hu2022lora}-based text-to-image fine-tuning. We use the VideoSwap~\cite{gu2024videoswap} dataset, which contains 30 videos sourced from Shutterstock and DAVIS~\cite{perazzi2016benchmark}, each paired with a reference image specifying the target identity across diverse categories (\eg, human, animal, object). Identity preservation is assessed using DINO-I~\cite{caron2021emerging} and CLIP-I~\cite{radford2021learning}, which capture structural and semantic alignment with the reference. As shown in Tab.~\ref{tbl:main_quantitative_table}, our method achieves the highest scores on both CLIP-I and DINO-I and all user study criteria, demonstrating its effectiveness in subject swapping. Our method ranks second in Flow Similarity and Warp Error because these metrics depend on precise pixel-level flow. However, our method still shows superior perceptual motion quality, as reflected in the user study results.

\input{tables/videoswap_quantitative}

\subsection{Analysis}

\begin{figure}[!t]
    \centering
    \vspace*{-2em}
    \hspace*{-2em}
    \includegraphics[width=0.5\textwidth]{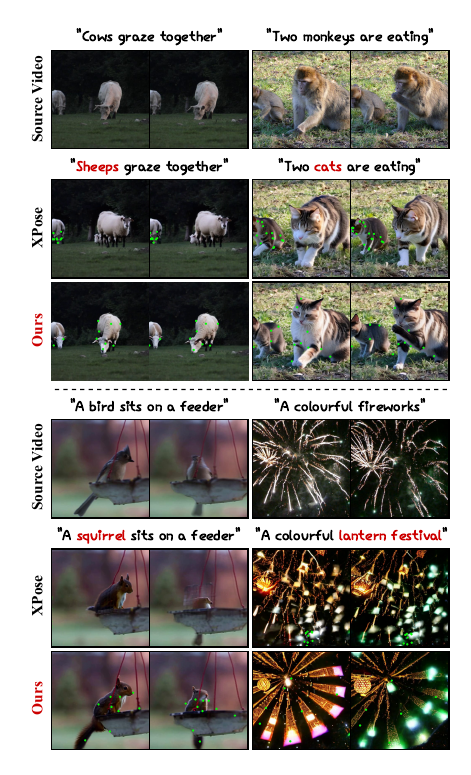}
    \vspace{-1.5\baselineskip}    
    \caption{Comparison with an existing open-world pose estimator~\cite{yang2024x}  across diverse video scenarios.}
    \label{fig:analysis}
\end{figure}

To demonstrate that our anchor tokens provide practical and robust points,
we compare results using anchor tokens against those from an open-world pose estimator~\cite{yang2024x}. While X-Pose requires a manually crafted prompt for each keypoint, anchor tokens can be applied without any manual input. Fig.~\ref{fig:analysis} illustrates the edited results, where the points are marked in green. Our anchor tokens robustly capture the overall motion in the scenes with multiple subjects (see 3rd row). They also remain effective even when a subject appears blurred due to rapid motion, or when the video depicts unstructured scenes without a focal subject (see 6th row), demonstrating the practicality and generality.

\subsection{Ablation Studies}

\noindent \textbf{Ablation on Selection Strategies.} A natural question arises: is sampling based on VDM's feature distance actually meaningful? To answer this, we compare our anchor tokens with those selected using alternative strategies: random sampling and sampling based on image feature~\cite{tang2023emergent} distance. Quantitative results show that our anchor tokens consistently outperform randomly sampled tokens ({\tiny \textsc{ViCLIP}} 0.234, {\tiny \textsc{Flow}} 0.777, {\tiny \textsc{Warp Err}} 0.370) and those selected using image feature ({\tiny \textsc{ViCLIP}} 0.234, {\tiny \textsc{Flow}} 0.776, {\tiny \textsc{Warp Err}} 0.365), achieving higher scores in most metrics ({\tiny \textsc{ViCLIP}} 0.257, {\tiny \textsc{Flow}} 0.711, {\tiny \textsc{Warp Err}} 0.336). This demonstrates that feature distance in the video diffusion model provides a more semantically meaningful and transferable signal for guiding motion. 

\medskip
\noindent \textbf{Ablation on Number of Anchor Tokens.} We also perform an ablation study by adjusting the number of anchor tokens, evaluating the impact of both sparser and denser selections. 
We adjust a hyperparameter $\tau$, which controls early stopping in the sampling procedure: a higher $\tau$ causes the sampling to stop earlier, as a newly selected token is considered redundant if its distance with any previously selected token falls below $\tau$. As shown in Fig.~\ref{fig:quantitative}, using fewer (sparser) anchor tokens tends to improve edit fidelity but leads to a drop in motion fidelity. In contrast, increasing the number of anchor tokens enhances motion fidelity but compromises edit quality. 
Based on these observations, we set $\tau=0.65$, which shows robust performance both quantitatively and qualitatively. 

%% file: tables/videoswap_quantitative.tex

\begin{table}[!t]
\footnotesize
\setlength{\tabcolsep}{0.1em}
\definecolor{Gray}{gray}{0.9}
\renewcommand{\arraystretch}{1.0}
\begin{tabularx}{\linewidth}{m{1.5cm} | Y Y Y Y | Y Y Y Y }
    \toprule
    \multirow{2}{*}{Method} & \multicolumn{4}{c|}{\textit{Automation Metrics}} & \multicolumn{4}{c}{\textit{User Study (Ratio)}} \\
    & \rule{0pt}{2ex} {\scriptsize DINO} & {\scriptsize CLIP} & {\scriptsize Flow} & {\scriptsize W-Err.} & {\scriptsize Edit} & {\scriptsize Motion} & {\scriptsize Temp.} & {\scriptsize Overall} \\
    \midrule\midrule
    MD & 0.285 & 0.633 & 0.275 & 0.576 & 0.148 & 0.158 & 0.179 & 0.128 \\
    FLATTEN & \underline{0.351} & \underline{0.676} & \textbf{0.794} & 0.303 & 0.306 & 0.446 & 0.397 & 0.373 \\
    TokenFlow & 0.243 & 0.665 & 0.352 & \textbf{0.180} & 0.059 & 0.095 & 0.087 & 0.073 \\
    \midrule
    \midrule
    \textbf{Ours} & \textbf{0.506} & \textbf{0.775} & \underline{0.550} & \underline{0.229} & \multicolumn{4}{c}{\textbf{v.s. Ours (1.000)}} \\
    \bottomrule
\end{tabularx}
\caption{Quantitative comparisons of VideoSwap~\cite{gu2024videoswap} datasets. \textbf{Bold} represents the best, and \underline{underline} represents the second best method.}
\vspace{-0.5\baselineskip}
\label{tbl:main_quantitative_table}
\end{table}

%% file: sections/05_conclusion.tex
\section{Conclusion}

In this paper, we introduced anchor tokens, a novel motion representation for video editing that captures essential motion patterns through a sparse set of point trajectories. They are fully automated and generalize across a wide range of subjects without requiring any manual input or human guidance.
By tracking meaningful token trajectories in the latent space of a video diffusion model, our approach enables accurate motion representation while allowing flexible subject replacement by adaptively relocating anchor tokens based on semantic correspondence. Extensive experiments demonstrate that our method achieves superior performance across diverse editing scenarios, offering a better trade-off between edit and motion fidelity. 





%% file: sections/06_appendix.tex
\clearpage
\appendix

\renewcommand{\thesection}{\Alph{section}}

\section*{Appendix}

\vspace{1em}
\hrule
\vspace{1em}

\input{sections/supplementary/implementation_details}
\input{sections/supplementary/additional_results}

\input{sections/supplementary/ablation_studies}
\input{sections/supplementary/human_eval_details}
\input{sections/supplementary/limitation}

\input{sections/supplementary/figures}

%% file: sections/supplementary/implementation_details.tex
\section{Implementation Details}

\textbf{Model Components and Hyperparameters.} In our experiment, we use the Chilloutmix\footnote{https://civitai.com/models/6424/chilloutmix} and incorporate the pre-trained motion module from AnimateDiff~\cite{guo2023animatediff} as the foundation for our method. Following \cite{gu2024videoswap}, we adopt lightweight MLPs as our anchor token encoder to balance performance with computational efficiency. The MLPs are fine-tuned for 50 -- 100 steps using a learning rate of $5 \times 10^{-5}$. During inference, we employ the DDIM sampler~\cite{song2021ddim} with 50 timesteps. We set the threshold $\tau$ for anchor tokens selection to 0.65. This means that if the distance between a newly selected anchor token and any previously selected anchor token drops below 0.65, we consider further sampling to be redundant and terminate the sampling procedure.

\medskip
\noindent \textbf{Feature Extraction.} To extract diffusion features for anchor token selection, we follow the general methodology described in prior works~\cite{tang2023emergent,xiao2024video,wang2024cove}. Specifically, we add noise corresponding to timestep $t = 261$ to each frame of the source video and feed the resulting noisy frames into the U-Net. We then extract features from an intermediate decoder layer. Following \cite{xiao2024video}, the extracted features $\tilde{g}$ are normalized by subtracting the mean across all frames: 
\begin{align}
 \tilde{f} = \tilde{g} - \frac{1}{N} \sum^{N}_{i=1} {\tilde{g}_i}
\end{align}
where $\tilde{g}_i$ denotes the feature of the $i$-th frame and $N$ is the total number of frames. During inference, we adopt the same process to extract diffusion features from the generated video, yielding features $f$ for anchor token alignment.

\medskip
\noindent \textbf{Experimental Setting.} All source videos are uniformly sampled to 16 frames and resized to either 512 $\times$ 512 or 448 $\times$ 768 resolution. For all baselines~\cite{zhang2023controlvideo,cong2023flatten,geyer2023tokenflow,zhao2024motiondirector,yatim2024space,yang2025videograin}, we use their official repository and default configuration. We use a depth estimator~\cite{birkl2023midas}, and an optical flow estimator~\cite{teed2020raft} for signal-based models. For X-Pose~\cite{yang2024x}, we follow the official repository’s prompts, using ``person'' for human videos and ``animal'' for animal videos while manually specifying prompts for other object types.
While we report results from a single run, we use a fixed random seed and deterministic sampling~\cite{song2021ddim} to ensure reproducibility~\cite{zhang2023emergence}.

\medskip
\noindent \textbf{Computing Infrastructure.} All experiments were conducted using a machine equipped with an NVIDIA A100 GPU (80GB VRAM). 
The environment is configured with Ubuntu 20.04 LTS, PyTorch 2.6.0, Python 3.10, and CUDA 12.4.
For anchor token selection, the memory cost includes 9GB for feature extraction and 9GB for optical flow estimation~\cite{teed2020raft}. During inference, anchor token alignment introduces no significant additional memory overhead beyond this baseline~\cite{guo2023animatediff, gu2024videoswap}, as anchor tokens operate in the latent space and similarity is computed over a reduced set of motion tokens.

%% file: sections/supplementary/additional_results.tex
\section{Additional Evaluation}

\subsection{Comparison with Pose Estimators}

We found that most existing pose estimators face challenges when applied directly to video editing tasks, due to the following reasons:
\begin{itemize}[leftmargin=*,noitemsep,topsep=0em]
    \item They are typically specialized for a particular domain (\eg, human or animal) and can only be applied to videos within their respective domains. 
    \item Their keypoints are often tightly coupled with semantic regions (\eg, body parts) rather than motion-related areas, which may result in suboptimal guidance for video editing tasks.
\end{itemize}
While the main paper focuses primarily on the first limitation, we focus on the second limitation in this section. We compare the performance of our anchor tokens against multiple off-the-shelf pose estimators~\cite{cao2019openpose,zhang2023clamp,yang2024x}, using two video sets for each domain. We select two subsets, human and animal, from our evaluation set, and additionally collect more videos from DAVIS~\cite{perazzi2016benchmark} and YouTube, resulting in 34 human videos and 15 animal videos. We apply X-Pose~\cite{yang2024x} using carefully curated text prompts appropriate for each domain. Tab.~\ref{tbl:supp_comparison} presents the video editing results when using our anchor tokens and, alternatively, keypoints from pose estimators. Using keypoints from OpenPose~\cite{cao2019openpose} for humans and CLAMP~\cite{zhang2023clamp} for animals yields higher motion fidelity scores than using keypoints from X-Pose~\cite{yang2024x}, an open-world pose estimator. This performance gain may stem from the advantage of fine-tuning on a specific domain. However, they often lead to degraded editing fidelity, possibly because their keypoints are concentrated on semantic parts of the subject, rather than being selected based on motion characteristics, causing the edited content to remain closely tied to the original subject. Meanwhile, our anchor tokens consistently achieve top or competitive scores, highlighting their robustness and effectiveness.
This indicates that, for video editing tasks, a straightforward, domain-agnostic selection of keypoints from a video diffusion model can achieve results  comparable to, or even better than, approaches that depend on external, domain-specific models and annotated data.

\input{tables/supp_comparison}

\subsection{Multiple Subject Editing} \label{subsec:multiple}

Existing video editing methods often struggle to handle multiple simultaneous motions~\cite{zhao2024motiondirector,jeong2023ground,huang2025videomage}, in particular, dense-signal-based methods tend to show degraded performance in multi-subject scenarios. Since these signals are continuous across all pixels, motion information from different subjects can interfere with one another due to occlusion or overlapping regions. In contrast, our anchor tokens, thanks to their sparse nature, are less affected by occlusion or overlapping regions, making our approach more robust in multi-subject scenarios. As shown in Fig.~\ref{fig:supp_qualitative}, our method effectively generates new subjects while preserving the original motion. In comparison, adaptation-based methods often fail to capture precise motion (blue boxes), while dense-signal-based methods tend to introduce artifacts, especially in regions where two subjects overlap (red boxes). For a quantitative evaluation, we collected 24 videos from Shutterstock and DAVIS~\cite{perazzi2016benchmark}, containing multiple subjects.
We introduce an additional metric to evaluate whether each method correctly generates all intended subjects in appropriate locations, since some methods over- or under-generate subjects and place them in irrelevant regions. We adopt Grounding-DINO~\cite{liu2024grounding} to detect the source and edited subjects in the video frames, treating the detections from the source video as ground truth. A prediction is considered correct if its IoU with a GT box exceeds 0.5. We compute precision (how many new subjects correctly match the source subjects) and recall (how many source subjects are successfully edited and appear in the edited video), and report their harmonic mean as the F1 score. Tab.~\ref{tbl:supp_multi_quantitative} shows that our method accurately generates multiple subjects, as reflected in both the F1 score and human evaluations.

\input{tables/supp_multi_quantitative}

%% file: tables/supp_comparison.tex
\begin{table}[!h]
\footnotesize
\setlength{\tabcolsep}{0.1em}
\definecolor{Gray}{gray}{0.9}
\renewcommand{\arraystretch}{1.0}
\begin{tabularx}{\linewidth}{m{1.5cm} m{1.5cm} | Y Y Y}
    \toprule
    & Method & ViCLIP $\uparrow$ & Flow Sim. $\uparrow$ & Warp Err. $\downarrow$ \\
    \midrule\midrule
    \multirow{3}{*}{Human} & OpenPose & 0.215 & 0.719 & 0.365 \\
    & X-Pose & 0.220 & 0.689 & 0.327 \\
    & Ours & \textbf{0.224} & \textbf{0.739} & \textbf{0.320} \\
    \midrule\midrule
    \multirow{3}{*}{Animal} & CLAMP & 0.234 & \textbf{0.813} & 0.341 \\
    & X-Pose & \textbf{0.235} & 0.738 & 0.691 \\
    & Ours & \textbf{0.235} & \underline{0.793} & \textbf{0.280} \\
    \bottomrule
\end{tabularx}
\caption{Comparison with video editing results using off-the-shelf pose estimators~\cite{cao2019openpose,zhang2023clamp,yang2024x}. \textbf{Bold} represents the best, and \underline{underline} represents the second best method.}
\label{tbl:supp_comparison}
\end{table}

%% file: tables/supp_multi_quantitative.tex

\begin{table}[!h]
\footnotesize
\setlength{\tabcolsep}{0.1em}
\definecolor{Gray}{gray}{0.9}
\renewcommand{\arraystretch}{1.0}
\begin{tabularx}{\linewidth}{m{2cm} | Y Y Y | Y Y Y Y }
    \toprule
    \multirow{2}{*}{Method} & \multicolumn{3}{c|}{\textit{Automation Metrics}} & \multicolumn{4}{c}{\textit{User Study (Ratio)}} \\
    & \rule{0pt}{2ex} {\scriptsize F1} & {\scriptsize Flow} & {\scriptsize W-Err.} & {\scriptsize Edit} & {\scriptsize Motion} & {\scriptsize Temp.} & {\scriptsize Overall} \\
    \midrule\midrule
    MD & 0.005 & 0.107 & 6.485 & 0.594 & 0.759 & 0.645 & 0.700 \\
    DMT & 0.241 & 0.535 & 3.668 & 0.378 & 0.378 & 0.417 & 0.378 \\
    FLATTEN & \underline{0.571} & \textbf{0.881} & 3.259 & 0.085 & 0.133 & 0.085 & 0.133 \\
    ControlVideo & 0.543 & 0.187 & 7.196 & 0.275 & 0.275 & 0.275 & 0.275 \\ 
    TokenFlow & 0.432 & 0.515 & \textbf{2.227} & 0.378 & 0.342 & 0.378 & 0.342  \\
    \midrule
    \midrule
    \textbf{Ours} & \textbf{0.601} & \underline{0.843} & \underline{2.644} & \multicolumn{4}{c}{\textbf{v.s. Ours (1.000)}} \\
    \bottomrule
\end{tabularx}
\vspace{0.2\baselineskip}
\caption{Multiple motion evaluation on videos from Shutterstock and DAVIS~\cite{perazzi2016benchmark}. \textbf{Bold} represents the best, and \underline{underline} represents the second best method.}
\label{tbl:supp_multi_quantitative}
\end{table}

%% file: sections/supplementary/ablation_studies.tex
\section{Ablation Studies}

\subsection{Ablation on Anchor Token Alignment} 

We conduct an ablation study to analyze the effect of anchor token alignment. As shown in Tab.\ref{tbl:supp_alignment}, removing anchor token alignment leads to a lower ViCLIP score. Without alignment, anchor tokens are injected at incorrect spatial locations relative to the new target subject, resulting in a generated subject with an awkward body shape that conforms to the structure of the subject in the source video, as illustrated in the second row of Fig.\ref{fig:supp_alignment}. On the other hand, when anchor token alignment is applied, the tokens are repositioned to match the structure of the new subject. This enables more accurate information injection and produces a more natural and well-formed result, as shown in the third row of Fig.~\ref{fig:supp_alignment}.

\input{tables/supp_alignment}

\subsection{Ablation on Trajectory Collection}

To obtain more robust and reliable candidates for anchor token selection, we first sample pixels from three keyframes (\(\{1, \lfloor \frac{N}{2} \rfloor, N\}\) for a source video with $N$ frames) and then track their trajectories across the source video. Therefore, we can capture more complete trajectories, including those that are partially occluded or move out-of-frame. In this section, we validate our trajectory collection strategy by comparing it with an alternative approach that samples pixels from a single frame and tracks their trajectories. Fig.~\ref{fig:supp_trajectory} shows that using a single keyframe can result in incomplete trajectory collection. Anchor tokens based on incomplete trajectories may become too dense or too sparse in specific frames, leading to lower edit fidelity ({\tiny \textsc{ViCLIP}} 0.243, 0.248, and 0.249 when using the first, middle, and last keyframe, respectively) compared to our method ({\tiny \textsc{ViCLIP}} 0.257).


\subsection{Ablation on Number of Anchor Tokens}

We conduct an ablation study to assess the impact of the number of anchor tokens. We adjust the hyperparameter $\tau$, which controls early stopping in the sampling procedure: a higher $\tau$ causes the sampling to stop earlier, as a newly selected token is considered redundant if its distance with any previously selected token falls below $\tau$. As shown in Fig.~\ref{fig:supp_num_anchor}, when the number of anchor tokens is too small (\eg, $\tau=0.9$), the edited video fails to accurately preserve the motion from the source, particularly subtle movements like changes in head direction. On the other hand, using too many anchor tokens (\eg, $\tau=0.5$) leads to overfitting to the subject's shape in the source video, resulting in artifacts such as elongated ears. We find that setting $\tau=0.65$ strikes a good balance, effectively preserving the original motion without being overly influenced by the subject's shape. The main paper presents quantitative results that follow a similar tendency.

%% file: tables/supp_alignment.tex
\begin{table}[!h]
\footnotesize
\setlength{\tabcolsep}{0.1em}
\definecolor{Gray}{gray}{0.9}
\renewcommand{\arraystretch}{1.0}
\begin{tabularx}{\linewidth}{m{3cm} | Y Y Y}
    \toprule
    Method & ViCLIP $\uparrow$ & Flow Sim. $\uparrow$ & Warp Err. $\downarrow$ \\
    \midrule\midrule
    \textit{wo} Anchor Alignment & 0.241 & \textbf{0.763} & 0.352 \\
    Ours & \textbf{0.257} & 0.711 & \textbf{0.336}  \\
    \bottomrule
\end{tabularx}
\caption{Ablation study on anchor token alignment.}
\vspace{-0.5\baselineskip}
\label{tbl:supp_alignment}
\end{table}

%% file: sections/supplementary/human_eval_details.tex
\begin{figure*}[ht!]
    \centering
    \includegraphics[width=\linewidth]{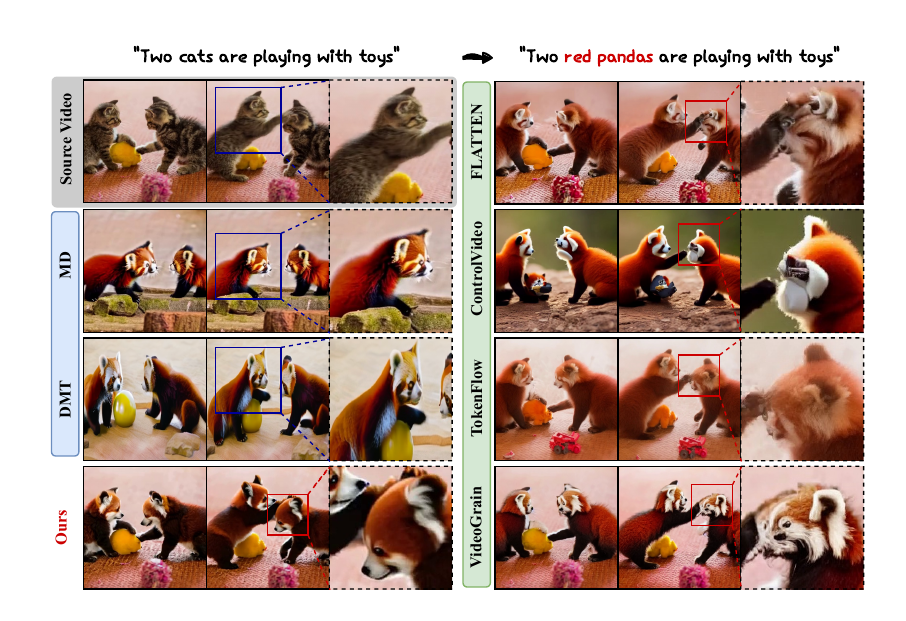}
    \caption{Qualitative comparison of multiple subjects editing results.}
    \label{fig:supp_qualitative}
\end{figure*}

\section{Human Evaluation Details}


We conduct user studies on Amazon Mechanical Turk (AMT) to compare our method with several baselines under two experimental settings, Customized Subject Swapping and Multiple Subject Editing. Each Human Intelligence Task (HIT) includes a source video, reference information (either images or a text instruction), and two edited results. Participants are asked to evaluate the two results based on the following four criteria:

\begin{itemize}[leftmargin=*,noitemsep,topsep=0em]
    \item \textbf{Edit fidelity}: Which video has a subject (or subjects) that better matches the reference information (e.g., reference images or text instruction)?
    \item \textbf{Motion fidelity}: Which video has subject motion that is more faithful to the source video?
    \item \textbf{Temporal consistency}: Which video appears smoother and more natural across frames?
    \item \textbf{Overall preference}: Overall, which video is better for the goal of subject replacement?
\end{itemize}

In the customized subject swapping setting, participants are shown a source video along with reference images of the target identity, as depicted in Fig.~\ref{fig:supp_amt_single}. In the multiple subject editing setting, participants are instead provided with a textual instruction (e.g., ``Replace the cars with jeeps''), as shown in Fig.~\ref{fig:supp_amt_multi}. Then, we collect responses for 52 HITs and 32 HITs, respectively. Each HIT consists of 10 video comparison set, including two screening questions to filter out unreliable responses. Each video pair are assigned to three independent participants. Workers are compensated at a rate of \$0.70 per HIT.


%% file: sections/supplementary/limitation.tex
\section{Limitation}

Anchor tokens rely on the semantic richness and stability of latent features. If the video diffusion model fails to capture meaningful motion-relevant features, anchor token selection may degrade. Additionally, since anchor token selection depends on latent feature trajectories, abrupt motion or heavy occlusion may lead to incomplete anchor candidates. Addressing these issues may require incorporating stronger temporal modeling or motion-aware priors in future work.

%% file: sections/supplementary/figures.tex
\begin{figure*}[t!]
    \centering
    \includegraphics[width=\linewidth]{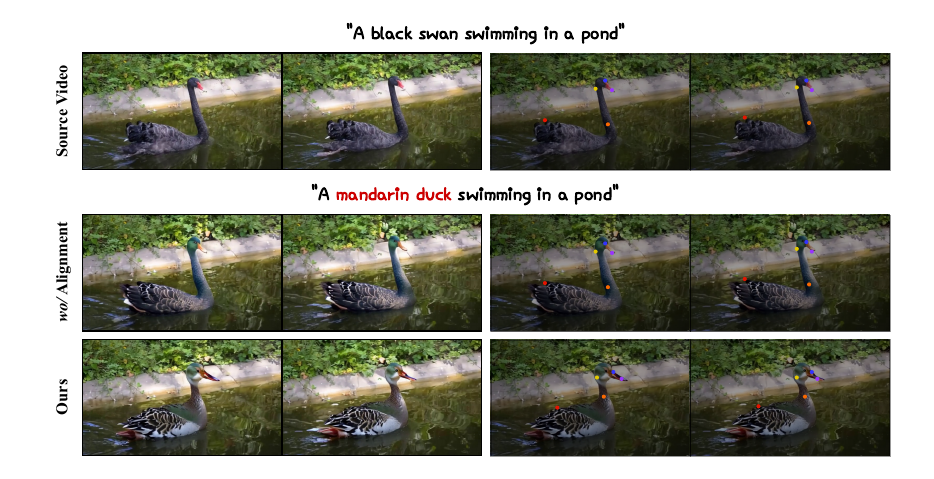}
    \caption{Effect of anchor token alignment on qualitative results.}
    \label{fig:supp_alignment}
\end{figure*}

\begin{figure*}[t!]
    \centering
    \includegraphics[width=\linewidth]{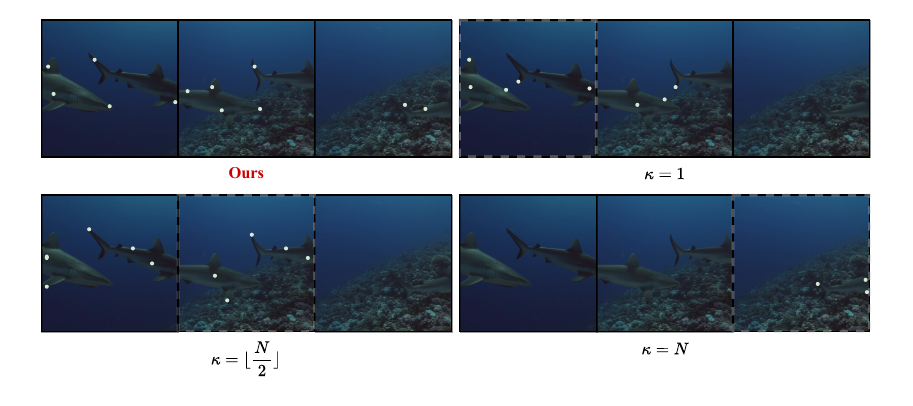}
    \vspace{-1em}
    \caption{Visualization of anchor tokens sampled from keyframes $\kappa$ and their trajectories where $N$ is the number of frames.}
    \label{fig:supp_trajectory}
\end{figure*}

\begin{figure*}[t!]
    \centering
    \includegraphics[width=\linewidth]{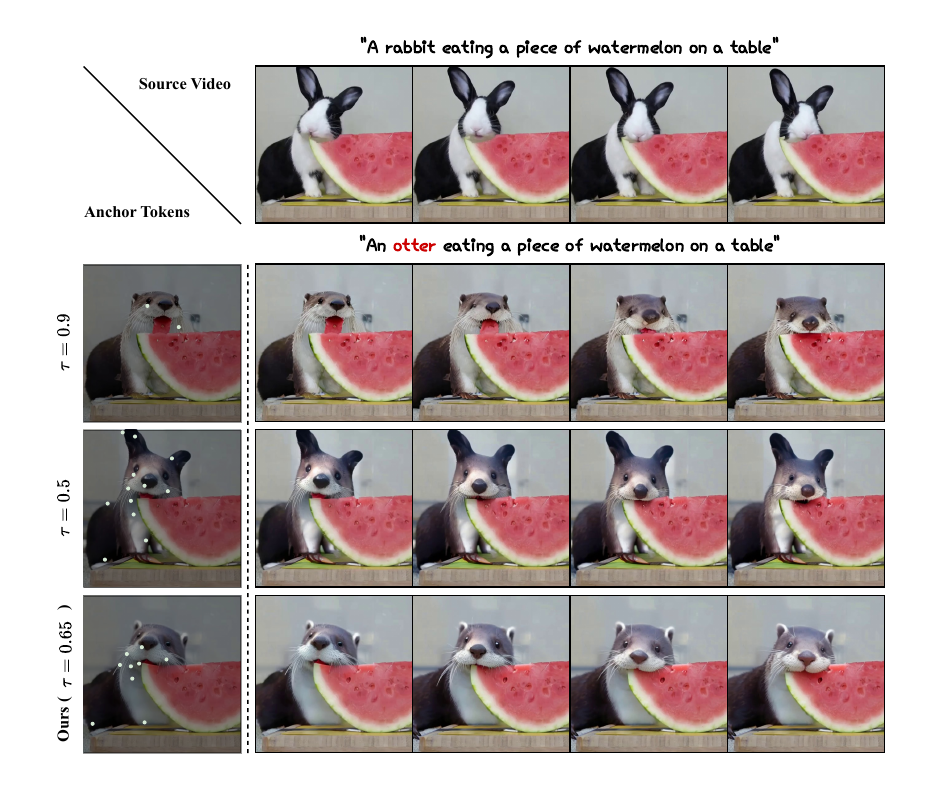}
    \caption{Effect of hyperparameter $\tau$ on qualitative results.}
    \label{fig:supp_num_anchor}
\end{figure*}

\begin{figure*}[t!]
    \centering
    \includegraphics[width=.7\linewidth]{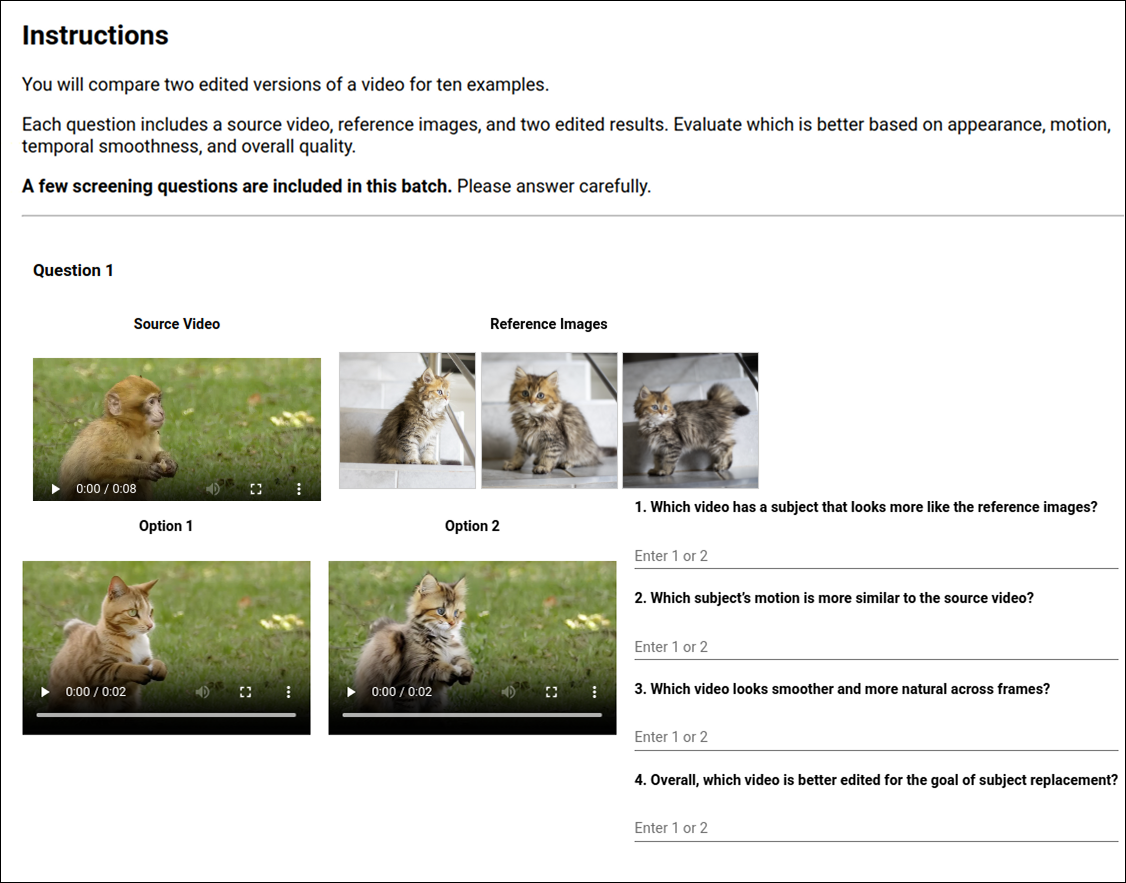}
    \vspace{1mm}
    \caption{AMT interface for customized subject swapping. Participants compare two edited videos with respect to the reference images and source video.}
    \label{fig:supp_amt_single}
\end{figure*}

\begin{figure*}[t!]
    \centering
    \includegraphics[width=.7\linewidth]{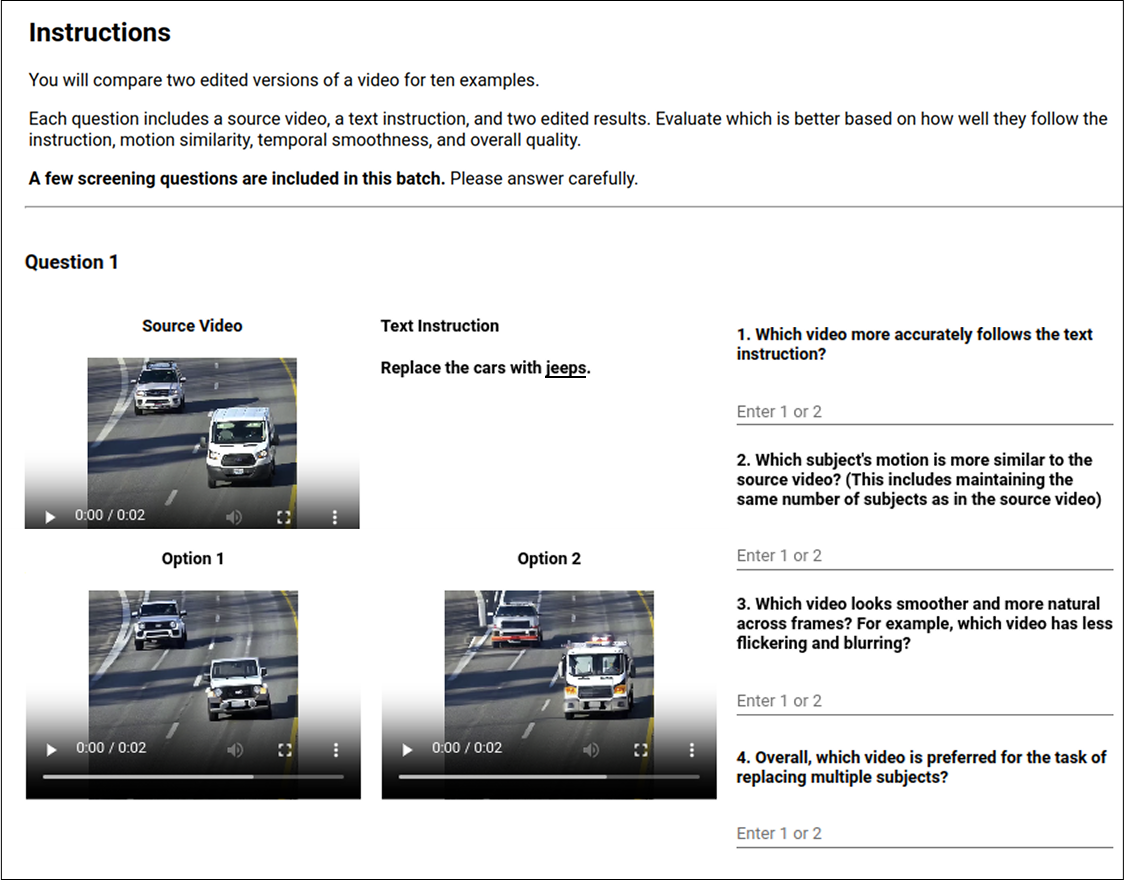}
    \vspace{1mm}
    \caption{AMT interface for multiple subject editing. Participants compare two edited videos based on a provided text instruction and the source video.}
    \label{fig:supp_amt_multi}
\end{figure*}